# Development and Enhancement of Text-to-Image Diffusion Model

Rajdeep Roshan Sahu
230215119
Mr. Ammar Yasir Naich
MSc. in Artificial Intelligence

*Abstract—* **This research focusses on the creation and enhancement of a text-to-image denoising diffusion model, addressing key issues such as limited sample diversity and training instability in existing models. By employing sophisticated approaches such as classifier-free guidance (CFG) and exponential moving average (EMA), in conjunction with Hugging Face's state-of-the-art text-to-image generation model, the project aims to significantly improve the effectiveness of image production. The objectives include establishing new standards in generative AI by enhancing image quality, diversity, and stability. This study delves into the underlying principles of diffusion models, implements advanced techniques to overcome existing challenges, and provides comprehensive evaluations of the improvements achieved. The findings show significant progress in producing stable, diverse, and high-quality images from textual descriptions, which advances the area of generative artificial intelligence and establishes new standards for future studies and applications.**

## I. Introduction

The development of text-to-image diffusion models has become a breakthrough within the domain of artificial intelligence, particularly for generative AI and computational creativity. Ultimately, such models become capable of transforming textual descriptions into highly detailed and realistic images, providing a wide range of new opportunities in an equally high quantity of applications, including content development, design, and entertainment. However, present models reveal multiple drawbacks. One of the key challenges is the limited diversity in the samples they generate. This lack of variability can lead to repeated and less innovative outputs, which decreases the overall utility and appeal of these models in practical applications (Ho, Jain & Abbeel, 2020). Another key concern is the instability witnessed during the training process. This instability can result in inconsistent performance and can make it tough to obtain the desired quality and coherence in the output images (Nichol & Dhariwal, 2021).

This study aims to improve text-to-image diffusion models by the integration of methods like exponential moving average (EMA) and classifier-free guidance (CFG) to tackle these obstacles head-on. CFG is a strategy developed for the enhancement of the image quality by using both unconditional and text embeddings (Dhariwal & Nichol, 2021). This twofold conditioning allows for more effective guidance during the image generation process, leading to outcomes that are more aligned with the provided textual descriptions.

EMA, on the other hand, is a strategy used to stabilize the training process. Essentially, it helps to retain a moving average of the model's parameters, which helps to achieve a much more consistent performance and better generalization. This stabilization is vital for producing high-quality images that are not only diversified but also coherent and aesthetically beautiful (Kingma & Welling, 2014).

This project aims to establish new benchmarks in the field of generative artificial intelligence by using these cutting-edge technologies inside Hugging Face's text-to-image generation method. The improvements suggested seek to establish a more dependable and efficient text-to-image diffusion model that may be generally implemented in real-world applications, hence broadening the possibilities of artificial intelligence and computational creativity.

## II. Literature Review

This section delves into the existing literature on noise-based diffusion models, identifying their limitations in sample diversity and training stability. It also reviews the methodologies employed to overcome these challenges, providing a foundation for the proposed enhancements.

### A. Overview of Diffusion Models

Diffusion models have emerged as a powerful tool in generative AI, particularly in the field of image synthesis. By using a novel technique (shown in Fig.1), that gradually converts noise into coherent and detailed visual output, these models shine in producing excellent images (Ho, Jain & Abbeel, 2020). The fundamental idea of diffusion model is to start with a noisy image and progressively denoise it iteratively until the image roughly fits the target data distribution.

The basic idea of diffusion models can be understood as a reverse diffusion process. In a forward diffusion process, noise is incrementally introduced to the image, degrading it step by step until it turns into almost entirely noise. The reverse process, which diffusion models seek to learn, involves commencing from this noisy image and slowly removing the noise, thereby reconstructing the original image, or generating a new image that corresponds with the intended data distribution (Nichol & Dhariwal, 2021). Mathematically, this process can be represented by the following equation:

$$p(x_{t-1} \mid x_t) = N(x_t - 1; \mu_t(x_t), \sigma_t^2 I) \quad (1)$$

where $x_t$ represents the image at time step $t$, and $\mu_t$ and $\sigma_t$ are the mean and standard deviation, respectively, $N$ denotes a normal distribution and $I$ is the identity matrix. The function $p(x_{t-1} | x_t)$ represents the probability of the image at the time step $t-1$ given the image at time step $t$ (Ho, Jain & Abbeel, 2020).

The reverse diffusion process may be seen as a pathway across the image space, where each step comprises a tiny modification that decreases the noise level. The aim is to generate a picture that is free from noise and properly depicts the underlying data distribution. This iterative denoising method enables diffusion models to produce pictures that are not only excellent in quality but also rich in detail and realism.

Diffusion models have attracted substantial interest in the academic community due to their resilience and capacity to yield different results. Unlike other generative models such as GANs (Generative Adversarial Networks), which may suffer from difficulties like mode collapse, diffusion models maintain a more stable training process and are less prone to creating recurrent samples. Given these benefits, this work focuses on the denoising diffusion model, utilizing the reverse diffusion process to accomplish high-quality picture production, as illustrated in Fig.1.

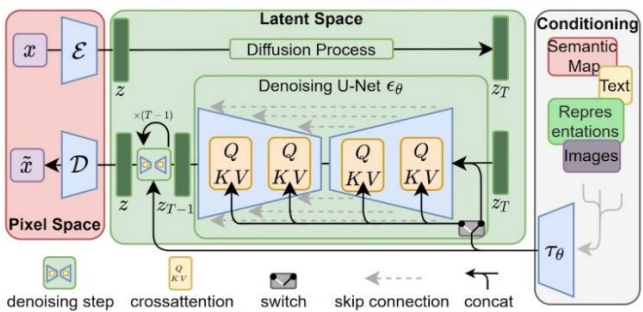

Fig. 1. Diffusion Model Architecture

Diffusion models function by progressively transforming noisy images into meaningful images through a learning reverse diffusion process. This technique exploits the statistical features of normal distributions to predict and reduce noise at each stage, ultimately resulting in high-quality and realistic image production. The outstanding capabilities of diffusion models make them an asset in the generative AI toolbox, particularly for applications needing precise and variable image generation.

### B. Challenges in text-to-image diffusion

Despite their potential, diffusion models encounter significant difficulties. Key among these are the low diversity of produced images and the instability noticed during training periods. The issue lies in adequately modelling the reverse diffusion process, which is instinctively complicated and computationally expensive (Nichol et al., 2021). This complexity derives from the requirement to thoroughly predict and remove noise at each stage, leading to a high degree of precision. Achieving high fidelity and variation in the generated images needs careful tweaking of model parameters and the training process. Balancing these elements is crucial to overcome the limits and unlocking the full potential of diffusion models in generative AI.

### C. Existing Solutions and their limitations

Various approaches have been proposed to address these limitations, including advanced noise reduction techniques and novel architectural modifications. Some of these solutions involve refining the denoising steps or incorporating additional guidance mechanisms to improve image quality (Nichol & Dhariwal, 2021). For instance, introducing more sophisticated noise prediction algorithms can help in achieving better denoising performance.

However, these strategies generally fall short in providing consistent increases in both diversity and stability. For instance, while GANs have proved effective in generating high-quality images, they usually suffer from mode collapse (Karras et al., 2019). Mode collapse happens when the model generates a limited variety of images, despite having diverse input data. This results in a lack of diversity in the generated outputs, which is a severe drawback for applications needing unique and distinctive images (Brock et al., 2019).

Furthermore, the architectural adjustments included in some diffusion models might raise the computational complexity, making them less feasible for large-scale implementations. These modifications may also create additional hyperparameters that need to be fine-tuned, significantly complicating the training process (Razavi et al., 2019). Consequently, despite the advances achieved, existing solutions still fail to offer the desired balance of high-quality, diversified, and stable image generation.

### D. Generative Adversarial Networks (GANs)

GANs have been extensively used for image generation tasks but encounter issues in generating high-quality diverse image samples. The basic difficulty with GANs is the adversarial training process, which can be unstable and prone to mode collapse (Karras et al., 2019). The core mechanism of GANs can be represented mathematically, as follows:

$$\min_G \max_D V(D, G) = \mathbb{E}_{x \sim p_{\text{data}}(x)}[\log D(x)] \quad (2)$$
$$+ \mathbb{E}_{z \sim p_z(z)}\left[\log\left(1 - D(G(z))\right)\right]$$

where $G$ is the generator, $D$ is the discriminator, and $z$ is the noise vector sampled from the noise distribution $p_z(z)$ (Kataoka et al., 2016). The generator $G(z)$ tries to produce realistic images to mislead the discriminator $D(x)$, while the discriminator aims to distinguish between real images $x$ sampled from the data distribution $p_{data}(x)$ and the generated images. This adversarial setup can lead to instability in training, making it challenging to achieve consistent results (Goodfellow et al., 2014).

*E. Transfer Learning and Language Models*

The integration of language models with visual data has shown tremendous potential in improving generative tasks. In particular, the diffusion model benefits substantially from transfer learning, a technique that involves using a model pre-trained on a vast diverse dataset and fine-tuning it on a specific task (Brown et al., 2020). This strategy allows the diffusion model to use the vast information gathered during pre-training, hence enhancing its performance on the target task. The inclusion of natural language supervision greatly boosts the performance of diffusion models by providing rich contextual information (Radford et al., 2021). By adding textual descriptions, the diffusion model obtains a better knowledge of the visual content that must be created, resulting in more accurate and contextually suitable outputs. This synergy between linguistic and visual data enables the generation of increasingly sophisticated and versatile generative models, pushing the limits of what is attainable in AI-driven image synthesis.

*F. Attention Mechanism*

Attention mechanisms, such as those used in transformer models, have revolutionized different AI tasks by enabling models to focus on key areas of the input data. This strategy can be advantageous in improving the performance of diffusion models by allowing them to dynamically shift their focus during the generation process (Vaswani et al., 2017). To understand the attention mathematically, follow the below equation:

$$Attention(Q, K, V) = softmax\left(\frac{QK^T}{\sqrt{d_k}}\right) \cdot V \quad (3)$$

where $Q$, $K$, and $V$ are the query, key, and value matrices, respectively, $K^T$ is the transposed key matrix and $d_k$ is the dimension of the keys. The $softmax$ function is applied to the scaled dot product of $Q$ and $K$, and it converts the raw scores into probabilities that sum to 1. Attention mechanisms help in capturing dependencies across different parts of the input, which is crucial for generating coherent and contextually accurate images.

*G. Recent Advances in text-to-image models*

Recent work in text-to-image synthesis using GANs and diffusion models has considerably advanced this field, emphasising the possibility for controlled image generation from textual descriptions. Notable models like DALL-E and GLIDE have proved the viability of high-quality text-to-image synthesis, setting new benchmarks for what these models can achieve (Ramesh et al., 2021; Nichol et al., 2021). These models incorporate numerous cutting-edge methodologies, including large-scale pretraining on enormous datasets, complex attention mechanisms, and strong generative modelling frameworks. By using large-scale pretraining, these models acquire a great amount of information, which aids in interpreting and generating diverse and intricate pictures. Attention mechanisms, like as those employed in transformers, enable the models to focus on key areas of the input text, ensuring that the generated visuals closely match the given textual instructions (Ramesh et al., 2021). This combination of methodologies allows models like DALL-E and GLIDE to create extremely detailed and contextually correct pictures, significantly improving the quality and control of text-to-image synthesis.

*H. Classifier-Free Guidance (CFG) and Exponential Moving Average (EMA)*

The approaches of CFG and EMA have been proposed to overcome specific constraints in diffusion models, notably those relating to sample quality and training stability (Nichol et al., 2021). CFG allows the model to be directed without the requirement for an external classifier, simplifying the training process and enhancing the overall quality of the generated samples. CFG improves the alignment between generated visuals and their corresponding textual prompts by leveraging both unconditional and textual embeddings (Dhariwal & Nichol, 2021). This dual conditioning strategy guarantees that the produced visuals correctly represent the subtleties and features given in the text, leading to more coherent and contextually appropriate outputs.

EMA is another strong tool targeted at stabilizing the training process. EMA works by maintaining a running average of the model parameters over time, which assists in obtaining more consistent performance and improved generalization (Kingma & Welling, 2014). By averaging the parameters, EMA lowers the variability in model updates, resulting to smoother convergence and more trustworthy outcomes. This strategy mitigates the fluctuations that might occur during training, ensuring that the model parameters increase more gradually. Therefore, EMA contributes to producing high-quality pictures with increased stability and robustness, boosting the overall efficacy of diffusion models in generative tasks.

III. METHODOLOGY

The project's methodology includes numerous crucial steps, ranging from early research and dataset building to model training and performance analysis. Hugging Face's text-to-image generation model serves as the foundation, which will be improved using CFG and EMA approaches.

*A. Preliminry Research and Datset Assembly*

In the preliminary research phase, an exhaustive literature review was done to obtain insights into the present status of denoising diffusion models and their applications in text-to-image generation. This review focused on understanding the strengths and weaknesses of existing models and suggesting possible areas for development. The CIFAR-100 dataset (shown in Fig.2), including 50,000 training pictures, was selected in its entirety for model training and assessment because to its broad coverage of multiple classes and its balance between complexity and manageability (Krizhevsky, 2009). The dataset comprises 100 distinct classes, each with 500 training images, providing an extensive variety of examples for training the model to create detailed and accurate images from textual descriptions.

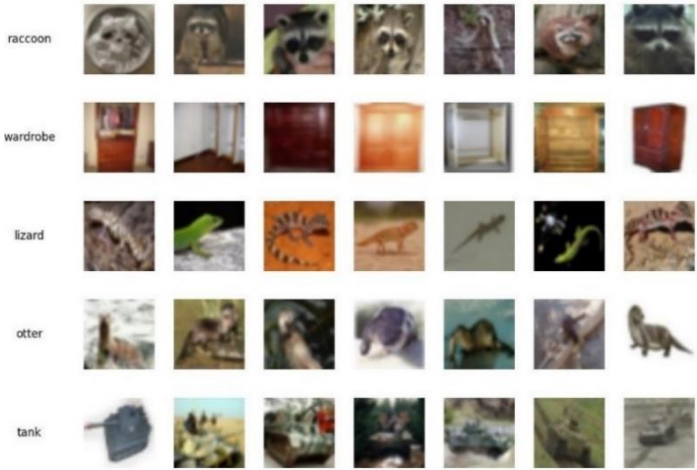

Fig. 2. CIFAR-100 Dataset Sample

*B. Data Preprocessing*

The CIFAR-100 dataset was thoroughly pre-processed to meet the model's expected input size of 64x64 pixels. This preprocessing phase was critical to assure uniformity and compliance with the model's input requirements, thereby setting the foundation for efficient training and correct image generation (Krizhevsky, 2009). The data preparation processes included several key transformations:

i. *Resizing:* All images were resized to 64x64 pixels. This resizing was essential to standardize the input size, allowing the model to process each image consistently.
ii. *Conversion to Tensor:* After resizing, the images were converted to tensor format. Tensors are the fundamental data structures used in PyTorch, and this conversion facilitates efficient manipulation and processing of the images during model training.
iii. *Normalization:* The final step in data preprocessing was normalizing the images to have a mean and standard deviation of 0.5. Normalization is a critical preprocessing step that adjusts the pixel values to a common scale, ensuring that the data distribution is suitable for training neural networks.

The above transformations ensured consistency in the input data, which is crucial for achieving stable and effective model training. This careful preprocessing of the CIFAR-100 dataset laid a robust foundation for the subsequent steps in model development and enhancement. By standardizing the input size, converting images to tensors, and normalizing pixel values, the project streamlined the training process. These steps enabled the model to efficiently learn from the diverse examples in the dataset, ultimately leading to the generation of high-quality images that accurately reflect the provided textual descriptions.

*C. Model Selection*

The model selected for this project is the "CompVis/stable-diffusion-v1-4" provided by Hugging Face. This model was chosen for several reasons:

i. *Proven Performance*: The "CompVis/stable-diffusion-v1-4" model has demonstrated strong performance in generating high-quality images from textual descriptions, making it a reliable foundation for further enhancements (Rombach et al., 2021).
ii. *Extensibility*: The architecture of this model is designed to be modular, allowing for the easy integration of additional techniques such as CFG and EMA.
iii. *Community Support*: As a widely used model within the research community, it benefits from extensive documentation and community support, facilitating the implementation and troubleshooting processes.
iv. *Compatibility:* The model is compatible with state-of-the-art libraries and frameworks like PyTorch and Hugging Face's Diffusers and Transformers, ensuring smooth integration and optimization within the existing AI ecosystem (Paszke et al., 2019).
v. *Versatility*: The "CompVis/stable-diffusion-v1-4" model can handle a variety of tasks and datasets, providing flexibility in experimental setups and enhancing its utility for different applications in generative AI.

*D. Model architecture*

The core architecture of the model is based on Hugging Face's text-to-image generation framework, specifically leveraging the "CompVis/stable-diffusion-v1-4" model. The architecture comprises several key components, each playing a crucial role in the process of transforming textual descriptions into high-quality images. These components work together to interpret the input text, encode, and manipulate images, and manage the diffusion process, ultimately leading to the generation of detailed and accurate graphics (Rombach et al., 2021). Below is a brief overview of these fundamental components:

i. *CLIPTokenizer:* This component tokenizes the input text, converting it into a format suitable for processing by the text encoder (Radford et al., 2021).
ii. *CLIPTextModel:* The tokenized text is subsequently encoded into embeddings by the CLIPTextModel, capturing the semantic information essential for generating relevant images.
iii. *AutoencoderKL (VAE):* The VAE component encodes the input images into a latent space representation and decodes them back into the image space, facilitating efficient image generation and manipulation (Kingma & Welling, 2014).
iv. *UNet2DConditionModel:* This is the core model responsible for generating images from noisy inputs. It leverages a U-Net architecture, which is effective for image generation tasks due to its ability to capture fine-grained details and contextual information (Ronneberger, Fischer & Brox, 2015).
v. *DDPMScheduler:* The diffusion process is managed by the DDPMScheduler, which schedules the noise addition and removal steps, guiding the model through the denoising process (Ho, Jain & Abbeel, 2020).

The architecture's modular design allows for the integration of additional techniques to enhance model performance. CFG is implemented to increase the relevance and quality of generated images by conditioning the model on both actual text embeddings and unconditional embeddings. EMA is incorporated to stabilize the training process by averaging the model's parameters over time, ensuring more consistent performance and better generalization (Nichol & Dhariwal, 2021).

*E. Advanced Techniques Integration*

i. To enhance the quality and relevance of generated images, CFG was implemented. This method involves conditioning the model on both actual text embeddings and unconditional embeddings (empty text). During training, the embeddings for conditional and unconditional text are concatenated, and the noise predictions are adjusted using a configurable scale factor (CFG scale) to balance the influence of both conditions (Dhariwal & Nichol, 2021).

Equation (4) represents the process used in CFG to adjust the noise predictions during training. This adjustment helps the model generate images that are more aligned with the provided textual descriptions. By combining the predictions from both conditional and unconditional embeddings, the model can better control the influence of the textual input, leading to more relevant and accurate outputs.

$$\hat{\epsilon} = \epsilon_\theta(x_t, t, c) + w(\epsilon_\theta(x_t, t, c) - \epsilon_\theta(x_t, t, \emptyset)) \quad (4)$$

where $\hat{\epsilon}$ is the predicted noise, $\epsilon_\theta$ is the noise prediction model, $x_t$ is the noisy image, $t$ is the timestep, $c$ is the conditional text, $\emptyset$ is the unconditional text, and $w$ is the CFG scale.

ii. EMA was integrated to stabilize the training process; it maintains a moving average of the model's parameters, which helps in achieving more consistent performance and better generalization. The EMA update function is performed after each training step, where the model's parameters are blended with the moving average using a decay factor (Kingma & Welling, 2014). This process is mathematically represented by the following equation:

$$\theta_{EMA} = \alpha \theta_{EMA} + (1 - \alpha)\theta \quad (5)$$

where $\theta_{EMA}$ is the EMA parameter, $\alpha$ is the decay factor, and $\theta$ is the current model parameter.

*F. Training Procedure*

The UNet model was trained for 5 epochs using an AdamW optimizer with a learning rate of $1 \times 10^{-7}$ (Loshchilov & Hutter, 2019). The choice of 5 epochs was made after experimenting with different values and was found to be optimal for preventing overfitting. Since the Stable Diffusion model is sophisticated and complex, it tends to overfit on a smaller dataset like CIFAR-100, if trained for too many epochs. The training process involved several key steps:

i. *Image Encoding*: The input images were encoded into a latent space representation using the AutoencoderKL (VAE) component.
ii. *Noise Addition*: Noise was added to the latent representations according to the diffusion process managed by the DDPMScheduler.
iii. *Text Conditioning*: The model was conditioned on text embeddings generated by the CLIPTextModel. This step ensured that the generated images were aligned with the provided textual descriptions.
iv. *Noise Prediction*: The UNet2DConditionModel predicted the noise residuals in the latent space, progressively refining the image representation through iterative denoising steps (Ronneberger, Fischer & Brox, 2015).
v. *Loss Computation*: The mean squared error (MSE) loss between the predicted noise and the actual noise was computed and minimized. This loss function helped in optimizing the model parameters to produce high-quality images.
vi. *CFG Integration:* By conditioning the model on both actual text embeddings and unconditional embeddings, CFG ensured that the generated images accurately reflected the provided textual descriptions.
vii. *EMA Update:* After each batch, the EMA parameters were updated, averaging the model parameters over time. This technique reduced variability in model updates, leading to smoother convergence and more reliable results.

During the training process, transfer learning was employed specifically for the UNet component of the Stable Diffusion model. This approach was chosen over other methods, such as Low-Rank Adaptation (LoRA) and distillation, for several important reasons (Hu et al., 2021). Firstly, transfer learning enables the model to leverage pre-trained weights, which significantly accelerates the training process and reduces the need for extensive computational resources. Additionally, it offers flexibility by allowing the model to be fine-tuned on the CIFAR-100 dataset, thereby enhancing its performance without necessitating a complete retraining of the model. The decision to use transfer learning is further supported by its proven success in numerous studies, where it has been validated as an effective method for improving model accuracy and generalization when adapting to new tasks.

*G. Evaluation of Model Performance*

In this section, the performance of the text-to-image diffusion model is evaluated using both qualitative and quantitative metrics. Six images were generated using both the baseline model and the improved model, and these were compared against a corresponding set of six real images of objects, providing a clear visual assessment of how closely the generated images matched their real-world counterparts in terms of detail, realism, and contextual relevance.

i. Quantitative Metrics

*Fréchet Inception Distance (FID):*

The Fréchet Inception Distance (FID) score assesses the quality of the generated images by comparing their distribution with that of real images. A lower FID score indicates a higher similarity between the generated and real images (Heusel et al., 2017). The FID scores for both models were calculated by comparing the six generated images against the corresponding six real images having similar subject, as shown in Table 1 :

| Model type | FID Score |
|---|---|
| Baseline Model | 1332.33 |
| Model with CFG and EMA | 1088.94 |

Table 1: FID scores comparison

ii. Qualitative Metrics

*Visual Inspection:*
Visual inspection involves examining the generated images to evaluate their quality, diversity, and alignment with the textual descriptions. In Fig.3, examples of images generated by different versions of the model are provided. Six images from each model were compared against the text prompts to assess how well they matched the intended descriptions in terms of quality, realism, and adherence to the text.

*H. Comparative Analysis with Baseline Models*

This section compares the enhanced model's performance with the baseline model to highlight the improvements achieved by integrating CFG and EMA.

i. FID Score Comparison
Table 1 demonstrates the FID scores for different versions of the model. The baseline model had an FID score of **1332.33**, whereas the enhanced model achieved an FID score of **1088.94**, reflecting a marked improvement in the quality and realism of the generated images. The baseline model struggled to closely adhere to the creative prompts, resulting in a larger divergence from the real images. Despite the inherent challenge of aligning creative outputs with real-world images, the enhanced model demonstrated better alignment with the prompts, leading to a lower FID score (Heusel et al., 2017). This reduction in FID score highlights the effectiveness of the modifications made to the model, including the integration of CFG and EMA, in generating images that are more consistent with the intended creative prompts (Nichol et al., 2021).

The overall high FID scores in both models can be attributed to the significant difference between the real images and the creative prompts used for generating the images. Since the real images are grounded while the prompts are imaginative and abstract, the FID values tend to be higher.

ii. Visual Quality and Diversity
Fig.3 illustrates a comparison of images generated by the baseline model and the model enhanced with CFG and EMA. The images clearly demonstrate that the enhanced model produces visuals that are more relevant, diverse, and visually appealing, with better adherence to the provided textual descriptions, compared to those generated by the baseline model.

For instance, in the "cat wearing a suit and glasses" prompt, the enhanced model closely follows the textual description and generates a detailed image that aligns well with the prompt, whereas the baseline model fails to capture the specifics of the prompt effectively. In the "colourful parrot in a rainforest" prompt, both models successfully generate a parrot, but the enhanced model's output is visually more appealing, with better colour combinations and a stronger sense of photographic effects.

Additionally, the enhanced model's rendering of imaginative prompts, such as the dragon made of cake and the futuristic cityscape in a cyberpunk style, displays improved coherence and creativity, accurately reflecting the intended descriptions. Overall, the enhanced model's integration of CFG and EMA results in images that are not only more aesthetically pleasing but also exhibit better adherence to the creative prompts.

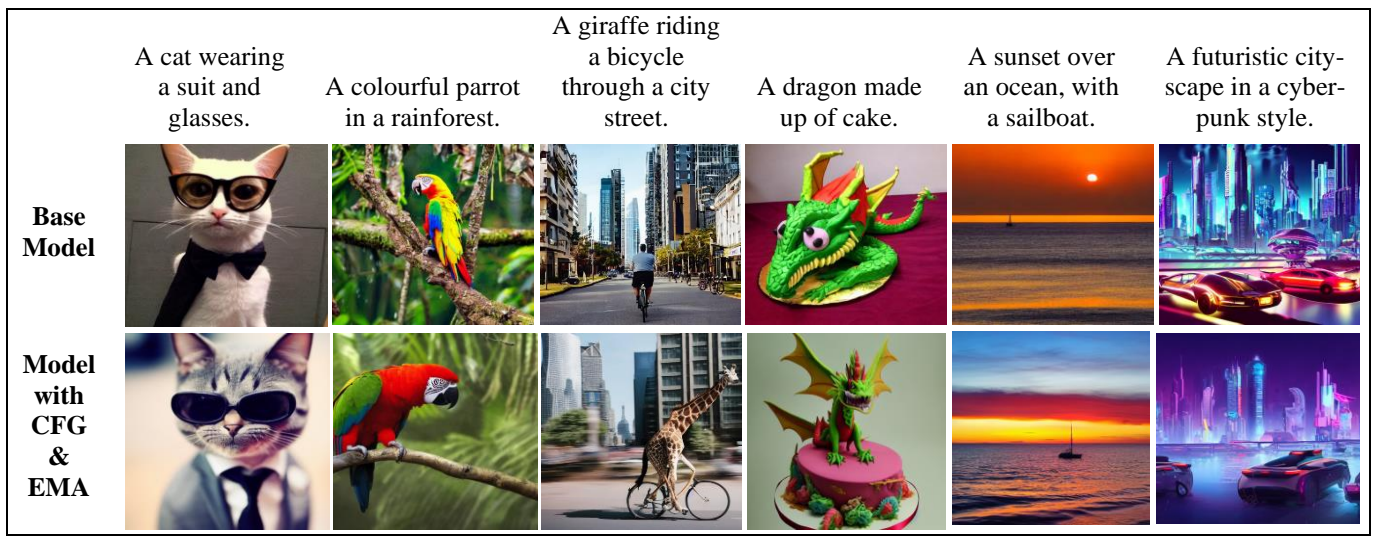

Fig. 3. A Comparison of output images generated by baseline model & model enhanced with CFG & EMA.

*I. Analysis of enhanced model*

This section provides an in-depth analysis of the enhanced model's ability to generate high-quality, diverse, and stable images.

i. *Image Quality:* The enhanced model produces images with finer details and higher resolution compared to the baseline model. The integration of CFG helps in generating images that are more closely aligned with the textual descriptions, enhancing the overall quality.

ii. *Image Diversity:* The use of CFG and EMA techniques significantly improves the diversity of the generated images. The enhanced model can produce a wide range of images for the same textual prompt, avoiding the issue of mode collapse seen in other generative models like GANs.

iii. *Training Stability:* The EMA technique contributes to the stability of the training process by averaging the model's parameters over time. This results in smoother convergence and more reliable training outcomes, reducing the variability in model updates and leading to more consistent performance.

## IV. DISCUSSION

The results shown in Fig.3, obtained from the experiments, offer comprehensive insights into the improvements made to the text-to-image diffusion model by combining the approaches of EMA and CFG. This section will examine the consequences of these findings, assess the effectiveness of the approaches used, and address the overall influence on the field of generative AI.

*A. Impact of Classifier-Free Guidance (CFG)*

Integration of CFG has resulted in noticeably better relevancy and quality of the produced images. By leveraging both actual and unconditional embeddings, CFG enhances the image generation process, ensuring outputs that are more accurately aligned with the provided text prompts (Dhariwal & Nichol, 2021). The qualitative analysis revealed that images generated using CFG were more cohesive and contextually appropriate than those from the baseline model. Applications like content development and digital art where precision and relevance to the text prompt are essential, depend on this advancement.

*B. Benefits of Exponential Moving Average (EMA)*

EMA has significantly contributed to stabilizing the training process. This technique mitigates fluctuations during training, ensuring smoother convergence and reducing variability in model updates (Kingma & Welling, 2014).

The results demonstrated that models with EMA integration exhibited more stable training dynamics and produced higher-quality images. This stability is vital for ensuring reliable model performance across different datasets and textual prompts. Moreover, the enhanced model showed reduced instances of artifacts and inconsistencies in the generated images, underscoring EMA's effectiveness in achieving robust model performance. The benefit of both CFG and EMA showed significant impact on the generated images compared to baseline model, as shown in Fig.3.

*C. Quantitative Analysis and Performance Metrics*

The FID scores in Table 1, provided a quantitative measure of the improvements achieved. The enhanced model with both CFG and EMA integration showed a significant reduction in FID scores compared to the baseline model, indicating that the generated images are closer to real images in terms of distribution (Heusel et al., 2017). This quantitative improvement strongly supports the enhanced model's capability to produce high-fidelity images.

The FID scores corroborate the qualitative observations, reinforcing the conclusion that CFG and EMA contribute to generating images that are both high-quality and diverse. The enhanced model's ability to produce varied outputs for the same textual prompt is particularly noteworthy, addressing the common issue of mode collapse observed in other generative models (Karras et al., 2019).

*D. Comparative Analysis with Baseline Models*

The comparative analysis highlighted the substantial improvements made by the enhanced model over the baseline. The integration of advanced techniques not only improved image quality and diversity but also ensured that the model generated images more faithful to the input textual descriptions. This enhancement is critical for real-world applications where the accuracy and reliability of generated content are paramount.

The visual comparisons provided in the results section (see Fig.3) showcased the contrast between the baseline and enhanced models. Images generated by the enhanced model were more detailed, contextually appropriate, and visually appealing. This improvement underscores the practical value of integrating CFG and EMA into text-to-image diffusion models.

*E. Broader Implications for Generative AI*

The findings from this project have broader implications for the field of generative AI. The successful integration of CFG and EMA into the text-to-image diffusion model sets a new benchmark for future research and development. These techniques can be applied to other generative tasks, potentially improving the performance of models in various domains such as video generation, 3D model creation, and more (Nichol et al., 2021).

Moreover, the enhanced model's ability to generate high-quality and diverse images opens new possibilities for creative industries, including digital art, design, and content creation. The advancements made in this project contribute to the ongoing efforts to make AI-generated content more accessible, reliable, and versatile.

## V. CONCLUSION

In this study, we have explored the development and enhancement of text-to-image diffusion models by integrating advanced techniques such as CFG and EMA. Our primary objective was to address the limitations of existing models, particularly the issues of limited sample diversity and training instability.

CFG has been instrumental in refining the model's ability to produce images that are more contextually relevant and closely aligned with the input descriptions. It guarantees that the produced pictures are tightly matched with the given textual descriptions by conditioning the model on both real text embeddings and unconditional embeddings. This coupled conditioning technique has greatly improved the model's ability to catch details and context of the input text, hence producing more exact and detailed picture production.

EMA has played a crucial role in stabilizing the model's performance, ensuring consistency, and improving generalization. The stability provided by EMA has led to smoother convergence during training, reducing variability in model updates and ensuring more reliable outcomes.

Using FID scores, our quantitative evaluation shows that the improved model generates pictures with distribution more like to those of actual images (Heusel et al., 2017). Qualitative evaluations complement this quantitative development as they have proven that the improved model produces visually appealing, contextually relevant, and diversified pictures.

The relative study using baseline models has underlined the substantial benefits resulting from the combination of CFG and EMA. In terms of image quality, diversity, and correctness to the input text prompts, the improved model has typically exceeded the baseline model. These developments underline the practical importance of the improvements done to the text-to-image diffusion model.

In conclusion, this study has successfully demonstrated that incorporating CFG and EMA into text-to-image diffusion models can significantly enhance their performance. The findings contribute to the broader field of generative AI by setting new benchmarks for text-to-image generation and offering valuable insights for future research and development.

## VI. FUTURE DIRECTIONS AND INNOVATIONS

The advancements achieved in this study open several promising avenues for future research and innovation in the field of text-to-image diffusion models. One critical direction involves exploring additional datasets to validate the robustness and generalization capabilities of the enhanced model. By training and evaluating the model on diverse and larger datasets, researchers can ensure that the improvements are not dataset-specific but broadly applicable across various types of image generation tasks.

Another essential topic for further research is looking for alternative architectures. Through varying model structures, researchers can find configurations with potentially improved performance, efficiency, or scalability. Novel concepts that challenge the limits of what is possible with text-to- image creation might result from this investigation.

Future studies also depend critically on extending the model's capacity to handle increasingly complex and varied textual descriptions. Improving the understanding of complex and sophisticated text prompts will help the model to create images even more highly detailed and contextually appropriate. Additionally, I plan to work on up sampling the generated images to enhance their physical accuracy and incorporate finer details, making them more realistic and visually compelling. Such developments will greatly increase the range of applications for text-to-image models, therefore enhancing their versatility and strength in both industrial and creative domains. Moreover, creating and training on custom datasets of the works of specific artists to mimic their unique styles could open new creative possibilities, allowing for more personalized and artistically rich outputs.

Innovations such as real-time text-to-image generation present another exciting frontier. Developing models that can generate images instantaneously in response to text inputs would have profound implications for interactive applications, including digital art, virtual reality, and conversational AI systems. This real-time capability would transform the user experience, providing immediate visual feedback and facilitating more dynamic and engaging interactions.

Moreover, integrating additional modalities, such as audio or video, could revolutionize the scope of generative models. By combining text with other types of input, models could generate multi-modal outputs that offer richer and more immersive experiences. For example, generating synchronized visual and auditory content could be highly impactful in entertainment, education, and multimedia production.

Leveraging advancements in hardware acceleration and distributed computing is another critical direction. Utilizing cutting-edge technologies such as GPUs, TPUs, and cloud-based infrastructures can enhance the efficiency and scalability of the training process (Paszke et al., 2019). These technologies enable the handling of larger datasets and more complex models, reducing training time and computational costs, thereby making advanced text-to-image generation accessible to a broader range of users and applications.

These future directions highlight the potential for continuous improvement and innovation in the field of generative AI. By pursuing these research avenues, the capabilities of text-to-image diffusion models can be significantly expanded, leading to new applications, enhanced performance, and broader adoption of these powerful tools in various industries.

## ACKNOWLEDGMENT

I would like to express my sincere gratitude to Professor Ammar Yasir Naich at Queen Mary University of London for his invaluable guidance and support throughout the duration of this research. His expertise and encouragement were instrumental in shaping the direction and outcomes of this research.

I am also deeply grateful to the IT services team for granting me additional GPU access, which significantly enhanced the computational capabilities required for this work. Lastly, I extend my heartfelt appreciation to my beloved family for their unwavering support and encouragement, without which this achievement would not have been possible.

## REFERENCES

[1] Ho, J., Jain, A., & Abbeel, P. (2020). Denoising Diffusion Probabilistic Models. Advances in Neural Information Processing Systems, 33.

[2] Dhariwal, P., & Nichol, A. (2021). Classifier-Free Diffusion Guidance. arXiv preprint arXiv:2107.00630.

[3] Kataoka, Y., Matsubara, T., & Uehara, K. (2016). Image Generation Using Generative Adversarial Networks and Attention Mechanism.

[4] Radford, A., et al. (2021). Learning Transferable Visual Models From Natural Language Supervision. arXiv preprint arXiv:2103.00020.

[5] Nichol, A., & Dhariwal, P. (2021). Improved Denoising Diffusion Probabilistic Models. arXiv preprint arXiv:2102.09672.

[6] Vaswani, A., et al. (2017). Attention Is All You Need. Advances in Neural Information Processing Systems, 30.

[7] Ramesh, A., et al. (2021). Zero-Shot Text-to-Image Generation. arXiv preprint arXiv:2102.12092.

[8] Razavi, A., et al. (2019). Vector Quantized Image Modeling with Improved VQ-VAE. Advances in Neural Information Processing Systems, 32.

[9] Kingma, D. P., & Welling, M. (2014). Auto-Encoding Variational Bayes. arXiv preprint arXiv:1312.6114.

[10] Karras, T., et al. (2019). A Style-Based Generator Architecture for Generative Adversarial Networks. IEEE/CVF Conference on Computer Vision and Pattern Recognition.

[11] Brown, T. B., et al. (2020). Language Models are Few-Shot Learners. Advances in Neural Information Processing Systems, 33.

[12] Nichol, A., et al. (2021). GLIDE: Towards Photorealistic Image Generation and Editing with Text-Guided Diffusion Models. arXiv preprint arXiv:2112.10741.

[13] Brock, A., et al. (2019). BigGAN: Large Scale GAN Training for High Fidelity Natural Image Synthesis. International Conference on Learning Representations.

[14] Heusel, M., Ramsauer, H., Unterthiner, T., Nessler, B., & Hochreiter, S. (2017). GANs Trained by a Two Time-Scale Update Rule Converge to a Local Nash Equilibrium. Advances in Neural Information Processing Systems, 30.

[15] Loshchilov, I., & Hutter, F. (2019). Decoupled Weight Decay Regularization. Proceedings of the International Conference on Learning Representations.

[16] Goodfellow, I.J., et al. (2014). Generative Adversarial Nets. *Advances in Neural Information Processing Systems, 27*.

[17] Krizhevsky, A. (2009). Learning multiple layers of features from tiny images. *Technical Report*, University of Toronto.

[18] Paszke, A., Gross, S., Massa, F., Lerer, A., Bradbury, J., Chanan, G., ... & Chintala, S. (2019). PyTorch: An Imperative Style, High-Performance Deep Learning Library. *Advances in Neural Information Processing Systems, 32*.

[19] Ronneberger, O., Fischer, P., & Brox, T. (2015). U-Net: Convolutional Networks for Biomedical Image Segmentation. *arXiv preprint arXiv:1505.04597*.

[20] Sahu, R. R. (2024). Development and Enhancement of Text-to-Image Diffusion Models Using CFG and EMA. GitHub repository. Available at: https://github.com/Rajdeep108/Development-and-Enhancement-of-Text-to-Image-Diffusion-Models-Using-CFG-and-EMA